\pdfoutput=1

\documentclass[11pt]{article}

\usepackage[final]{ACL2023}

\usepackage{times}
\usepackage{latexsym}

\usepackage[T1]{fontenc}

\usepackage[utf8]{inputenc}

\usepackage{microtype}
\usepackage{cite}
\usepackage{amsmath,amssymb,amsfonts}

\usepackage{graphicx}
\usepackage{textcomp}
\usepackage{xcolor}

\usepackage{makecell}
\usepackage[ruled,vlined,linesnumbered]{algorithm2e}

\usepackage{booktabs,tabularx}
\usepackage{url}
\usepackage{latexsym}
\usepackage[subfigure]{graphfig}
\usepackage{multirow, booktabs}
\usepackage{footnote}
\usepackage{CJKutf8}

\usepackage{caption}
\usepackage{subcaption}
\usepackage{listings}

\usepackage{enumitem}

\usepackage{stfloats}
\usepackage{setspace}
\usepackage{pifont}
\usepackage{tcolorbox}
\usepackage{soul}
\usepackage{inconsolata}
\usepackage{makecell}
\usepackage{pifont}

\newcommand{\method}{SQL-Synth}

\tcbset{
  examplebox/.style={
    colback=gray!10, 
    colframe=gray!50, 
    coltitle=white, 
    fonttitle=\bfseries, 
    title={Example}, 
    boxrule=0.5mm, 
    rounded corners, 
    width=1\textwidth, 
    sharp corners=downhill, 
  }
}

\title{Beyond SELECT: A Comprehensive Taxonomy-Guided Benchmark for Real-World Text-to-SQL Translation}

\definecolor{verylightgray}{gray}{0.9}
\author{
Hao Wang\textsuperscript{1},
Yuanfeng Song\textsuperscript{1},
Xiaoming Yin\textsuperscript{1},
Xing Chen\textsuperscript{1} \\
\textsuperscript{1}ByteDance, China \\
}

\begin{document}
\maketitle
\begin{abstract}
Text-to-SQL datasets are essential for training and evaluating text-to-SQL models, but existing datasets often suffer from limited coverage and fail to capture the diversity of real-world applications. To address this, we propose a novel taxonomy for text-to-SQL classification based on dimensions including core intents, statement types, syntax structures, and key actions. Using this taxonomy, we evaluate widely used public text-to-SQL datasets (e.g., Spider and Bird) and reveal limitations in their coverage and diversity. 
We then introduce a taxonomy-guided dataset synthesis pipeline, yielding a new dataset named \textit{\method{}}. This approach combines the taxonomy with Large Language Models (LLMs) to ensure the dataset reflects the breadth and complexity of real-world text-to-SQL applications. 
Extensive analysis and experimental results validate the effectiveness of our taxonomy, as \textit{\method{}} exhibits greater diversity and coverage compared to existing benchmarks. 
Moreover, we uncover that existing LLMs typically fall short in adequately capturing the full range of scenarios, resulting in limited performance on \textit{\method{}}. However, fine-tuning can substantially improve their performance in these scenarios. 
The proposed taxonomy has significant potential impact, as it not only enables comprehensive analysis of datasets and the performance of different LLMs, but also guides the construction of training data for LLMs.

\end{abstract}

\section{Introduction}

In the field of database systems, SQL serves as a universal language for data manipulation and querying across diverse applications \citep{yu2018spider,song2024speech,zhang2024automatic}. Text-to-SQL, a widely used technique in real-world applications, translates natural language (NL) to structural SQL query, enabling users to interact with database system without expert knowledge. 
The design and optimization of text-to-SQL systems depend heavily on high-quality text-to-SQL datasets to ensure robustness and effectiveness. However, existing text-to-SQL datasets often suffer from limited coverage and diversity. Many datasets such as Spider \citep{yu2018spider} and Bird \citep{li2024can} are tailored to specific scenarios or applications and usually emphasize ``Select'' operations, failing to capture the full spectrum of user questions and SQL queries encountered in real-world scenarios. This shortcoming raises concerns about the ability of these datasets to effectively evaluate and enhance models applicability and capability. 

To tackle the aforementioned challenges, we propose a novel taxonomy for text-to-SQL data classification. This taxonomy offers a systematic framework that categorizes data across multiple dimensions, including core intent, statement type, syntax structure, and key action. Incorporating this taxonomy, we evaluate the coverage and diversity of widely used datasets including Spider and Bird. Our analysis uncovers their significant gaps between these datasets and real-world distributions, highlighting their limited coverage within our taxonomy and their lack of diversity in capturing real-world scenarios. 

To bridge these gaps, we propose a taxonomy-guided dataset synthesis pipeline, resulting in a new dataset named \textbf{\method{}}. This pipeline integrates the carefully designed taxonomy with Large Language Models (LLMs) to ensure the resulting dataset reflects the coverage and diversity of real-world text-to-SQL application. 
Specifically, the pipeline begins with generating valid taxonomy combinations to guide subsequent synthesis process. Then it enhances tables from WiKiSQL to construct more complex and cross-domain databases, enabling richer scenarios. Next, it produces high-quality seed data, which not only serves as templates for diversity expansion but also ensures coverage of the taxonomy. Finally, the pipeline employs a dual-path diversity expansion mechanism to expand dataset's diversity leveraging the enhanced databases.

Extensive analysis shows that \textbf{\method{}} achieves markedly broader coverage and richer diversity than existing benchmarks.
Moreover, we found that existing LLMs like Qwen2.5~\citep{qwen2025qwen25technicalreport}, Qwen2.5 Coder~\citep{hui2024qwen25codertechnicalreport}, and Granite3.1~\citep{mishra2024granitecodemodelsfamily} consistently fail to capture the full spectrum of user questions and SQL queries, yielding limited performance on SQL-Synth; nevertheless, fine-tuning can substantially improve their effectiveness in this particular scenario. 

In a nutshell, our contributions include: \begin{itemize}
\item We proposed a comprehensive taxonomy to reflect the coverage and diversity of text-to-SQL datasets.
\item We conducted an extensive analysis of existing text-to-SQL datasets using our proposed taxonomy and identified their shortcomings, such as limited coverage of complex queries found in real-world applications. 
\item  We designed a taxonomy-guided dataset synthesizing pipeline and introduced a new dataset called SQL-Synth. Extensive experimental analysis validates the rationales of the proposed taxonomy.
\end{itemize}

\section{Related Work}

\subsection{Text-to-SQL Models}
Text2SQL is a pivotal research topic at the intersection of natural language processing and databases systems. Its goal is to automatically translate human natural language questions into SQL statements, thereby enabling an effective Natural Language Interfaces (NLIs) to relational databases. The history of existing Text2SQL models is extensive and has gone through several stages, including rule-based \citep{Baik_2019, 10.1145/3133887}, neural-based \citep{hui2021dynamic, choi-etal-2021-ryansql, dou2022unisarunifiedstructureawareautoregressive}, LLM-based \citep{gu2023interleavingpretrainedlanguagemodels, guo2023retrievalaugmentedgpt35basedtexttosqlframework, sun2024sqlpalmimprovedlargelanguage}, and agent-based approaches \citep{cen2025sqlfixagent, deng2025reforcetexttosqlagentselfrefinement, xie2024magsqlmultiagentgenerativeapproach}. 
Among these, some notable neural-based works include IRNet \citep{guo2019towards}, while LLM-based methods encompass Din-sql \citep{pourreza2023din} and Purple \citep{ren2024purple}. Additionally, agent-based approaches feature examples such as Spider-Agent \citep{leispider} and ReFoRCE~\citep{deng2025reforcetexttosqlagentselfrefinement}.

\subsection{Existing Text-to-SQL Datasets}

In parallel with progress in models, the text-to-SQL community has released a variety of benchmarks. The most widely used are WikiSQL \citep{wikisql}, Spider \citep{yu2018spider}, and Bird \citep{li2024can}. Besides these general-domain datasets, there are also many domain-specific datasets, such as FinSQL in the financial domain \citep{zhang2024finsqlmodelagnosticllmsbasedtexttosql}, BookSQL in the accounting domain \citep{kumar2024booksql}, and ScienceBenchmark in the science domain \citep{zhang2023sciencebenchmark}.
In addition to English datasets, there are also datasets in other languages, such as CSpider \citep{min2019pilotstudychinesesql} in Chinese and MultiSpider \citep{dou2022multispiderbenchmarkingmultilingualtexttosql} in multiple languages. Moreover, there are datasets focused on multi-turn dialogues, like CoSQL \citep{yu-etal-2019-cosql}, and those emphasizing robustness, such as Spider-Rob \citep{chang2023drspiderdiagnosticevaluationbenchmark}. Even so, the coverage of these cases in existing datasets remains quite limited. Most of these datasets primarily focus on `Select' operations. Our work is the first to comprehensively address a broader range of SQL operations, thereby providing a more holistic approach to text-to-SQL tasks.

\begin{figure*}[th!]
  \centering
  \includegraphics[width=0.98\textwidth]{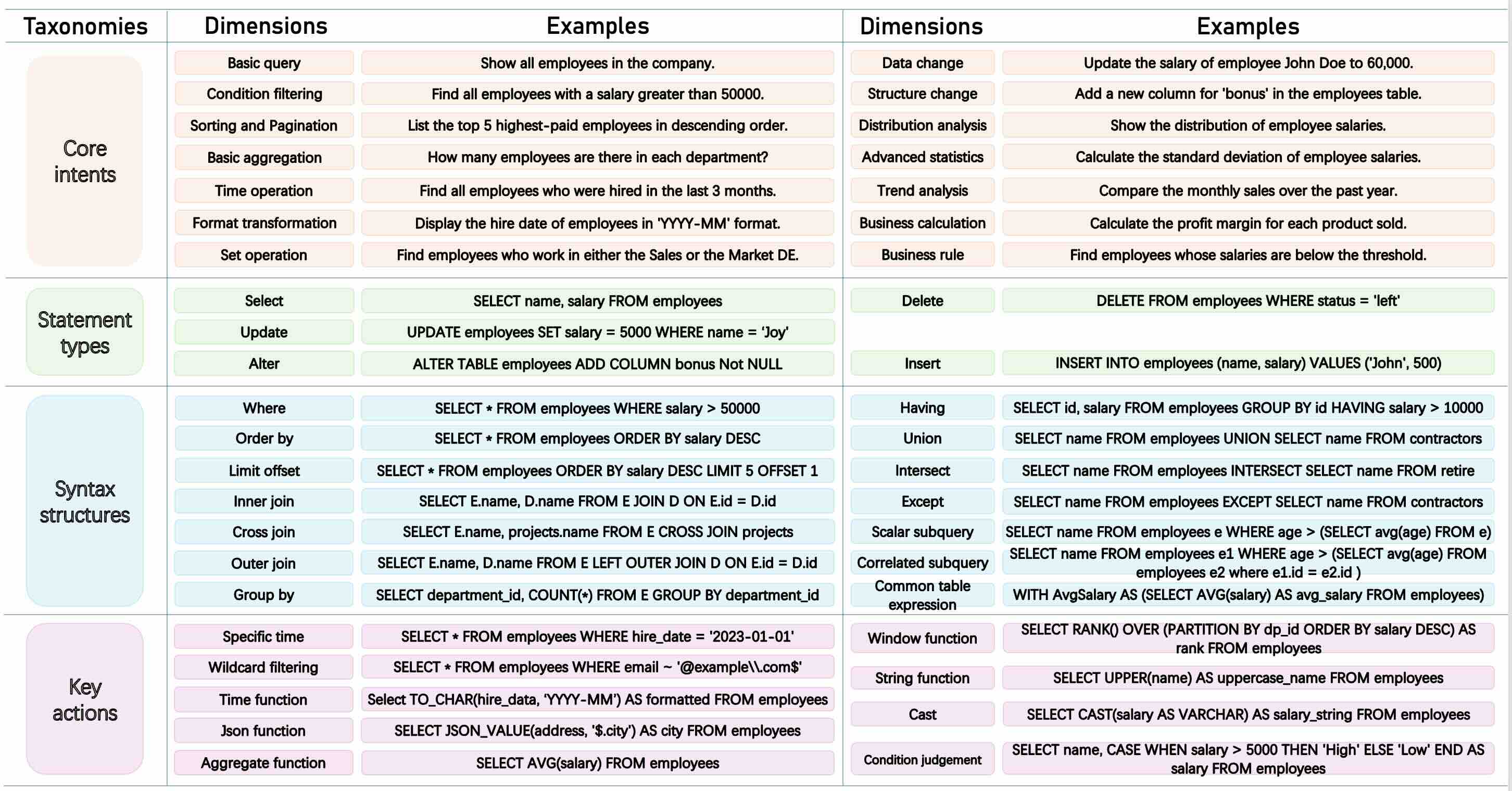}
  \vspace{-5pt}
  \caption{The proposed taxonomy for text-to-SQL. The taxonomy consists of four main dimensions, including Core intents, Statement types, Syntax structures, and Key actions. Core intents focus on the underlying purpose of the user question. In contrast, Statement types, Syntax structures, and Key actions emphasize the specific implementation details from the perspective of the resulting SQL query.}
  \vspace{-12pt}
  \label{fig:tax}
\end{figure*}

\section{Taxonomy for Text-to-SQL Types}

\subsection{Overview}
The task of designing a taxonomy for user question and SQL query requires a clear and structured understanding of common text-to-SQL scenarios. Based on the extensive analysis of text-to-SQL business scenarios, we propose a comprehensive two-level taxonomy to categorize text-SQL pairs.

Specifically, we categorize text-SQL pairs from two complementary perspectives: user questions and SQL queries. This dual approach is based on the observation that SQL queries may not always fully align with the intent of the corresponding user questions, and user questions always contain the complete and unhidden user intents. By examining both perspectives, we ensure comprehensive coverage of real-world scenarios in text-to-SQL. 
Therefore, as shown in Figure~\ref{fig:tax}, we categorize Text-SQL pairs along four main dimensions including Core intents, Statement types, Syntax structures, and Key actions. Each dimension is further decomposed into fine-grained subcategories, allowing for a more detailed and comprehensive coverage of real scenarios. 

\subsection{Two-Level Taxonomy}

\subsubsection{Core intents}
Core intents represent the fundamental purposes or objectives that a user aims to achieve when querying a database. Since the process of generating SQL queries often abstracts and obscures the original user intents, we categorize these intents based on the nature of the user's questions. These intents can be broadly classified into the following categories:

\noindent\textbf{Basic query:} It indicates user's request to retrieve data straight from a database, without transformations, filtering, or complex conditions.

\noindent\textbf{Condition filtering:} It refers to filter query results applying logical operators, comparison expressions, or pattern matching.

\noindent\textbf{Sorting and Pagination:} It involves sorting query results and dividing them into subsets for display and navigation.

\noindent\textbf{Basic aggregation:} It refers to the process of calculating statistics or combining data, such as COUNT or other simple aggregation functions.

\noindent\textbf{Time operation:} It involves handling temporal data, such as handling time values, calculating intervals, or performing operations on date fields.

\noindent\textbf{Format transformation:} It refers to converting data types or restructuring the format of the outputs to meet specific requirements.

\noindent\textbf{Set operation:} It involves combining results from multiple queries using set-based logic.

\noindent\textbf{Data change:} It refers to operations that alter the content of a database, such as inserting, updating, or deleting records.

\noindent\textbf{Structure change:} It involves modifying the database schema, such as altering table structures or updating column definitions.

\noindent\textbf{Distribution analysis:} It refers to analyzing the spread, frequency, and statistical dispersion of data to understand its distribution.

\noindent\textbf{Advanced statistics:} It involves performing complex mathematical computations and statistical analyses to derive deeper insights instead of basic aggregations.

\noindent\textbf{Trend analysis:} It focuses on identifying patterns, changes, and movements within data over time, enabling insights into temporal dynamics.

\noindent\textbf{Business calculation:} It involves computing domain-specific metrics or key performance indicators (KPIs) to support business analysis and decision-making.

\noindent\textbf{Business rule:} It refers to applying organizational logic, policies, or conditional constraints to ensure data processing aligns with specific business requirements.

\subsubsection{Statement types}
Statement types represent the categories or purposes of SQL queries being executed. While existing datasets primarily focus on SELECT operations, real-world scenarios often involve a broader range of operations, such as updating, altering, and deleting data. To address this gap, we classify statement types into the following categories: \textbf{Select}, \textbf{Update}, \textbf{Alter}, \textbf{Delete}, and \textbf{Insert}.

\begin{figure*}[th!]
  \centering
  \includegraphics[width=1\textwidth]{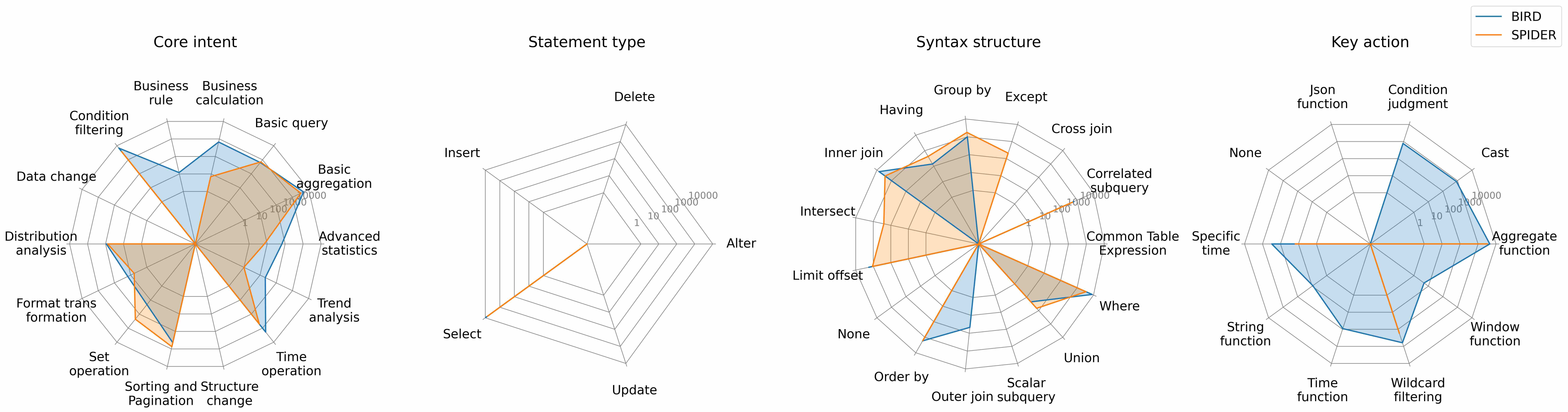}
  \caption{Taxonomic distribution of public Spider and Bird datasets.}
  \label{fig:Taxonomy distribution of Spider and Bird}
  \vspace{-4mm}
\end{figure*}

\subsubsection{Syntax structures}
Syntax structures represent the rules and components used to construct SQL queries, which are decisive for performing specific operations. 
Specifically, they encompass a variety of elements, including \textbf{Where} for conditional filtering, \textbf{Order by} for sorting, and \textbf{Limit offset} for display restrictions. Additionally, they include different types of joins, namely \textbf{Inner join}, \textbf{Cross join}, and \textbf{Outer join}, for combining tables, as well as grouping and aggregation mechanism with \textbf{Group by} and \textbf{Having}. SQL query can be merged or compared using set operations such as \textbf{Union}, \textbf{Intersect}, and \textbf{Except}. \textbf{Scalar subquery} and \textbf{Correlated subquery} allow for nested query logic, while \textbf{Common Table Expression} facilitate modular query design. Together, these syntax structures form the skeletons of SQL queries, enabling diverse and complex data manipulations. Please refer to Appendix~\ref{Taxonomy Details} for more detailed definitions.

\subsubsection{Key actions}
Key actions represent specific operations or functionalities performed within a SQL query to manipulate, filter, transform, or analyze data. For instance, \textbf{Specific time} is included when hardcoded temporal data is used, and \textbf{Wildcard filtering} is employed with the LIKE operator to match patterns in column values. 
Also, functions play a key role, such as \textbf{Time function} for manipulating temporal data, \textbf{Json function} for interacting with json structures, and \textbf{Window function} paired with $\mathrm{OVER()}$ for windowed calculations. 
Additionally, \textbf{String function} enable non-regex string manipulations, while the \textbf{Cast} allows explicit data type conversions. Queries may also utilize \textbf{Condition judgement} to apply conditional logic for determining output values, and \textbf{Aggregate function} to perform calculations across multiple rows of data. Together, these key actions enhance the flexibility and functionality of SQL qeries. Please refer to Appendix~\ref{Taxonomy Details} for more detailed definitions.

\subsection{Analysis of Existing Text-to-SQL Datasets}

As mentioned before, existing text-to-SQL datasets often suffer from limitations in coverage and diversity. Therefore, in this section, we analyze the coverages and diversity of two notable datasets on our taxonomy, namely Spider and Bird. 

Figure~\ref{fig:Taxonomy distribution of Spider and Bird} illustrates the coverage of the Spider dataset and Bird dataset across various dimensions. As shown, in terms of statement types, Spider exclusively includes the \textbf{Select} type. Regarding syntax structures, the dataset covers approximately $70\%$ of the possible structure types. For key actions, Spider is limited to only three action types, and for core intents, it encompasses eleven intent types. Additionally, the distribution across each taxonomy is noticeably uneven, with certain types being heavily represented while others are sparsely covered. 
Similarly, the Bird dataset struggles with incomplete taxonomy coverage and faces the challenge of uneven distribution across its coverage.

In summary, the taxonomy coverage and diversity of both the Spider and Bird datasets are insufficient to effectively capture the complexity and variety of common real-world scenarios. This limitation hinders the scalability and broader applicability of these datasets.

\section{Taxonomy-guided Synthesis of the SQL-Synth Dataset}
In this section, we first introduce how SQL-Synth is curated based on the popular text-to-SQL benchmark Spider using our proposed taxonomy, and then overview the composition and statistics of SQL-Synth.

\begin{figure*}[th!]
  \centering
  \includegraphics[width=1\textwidth]{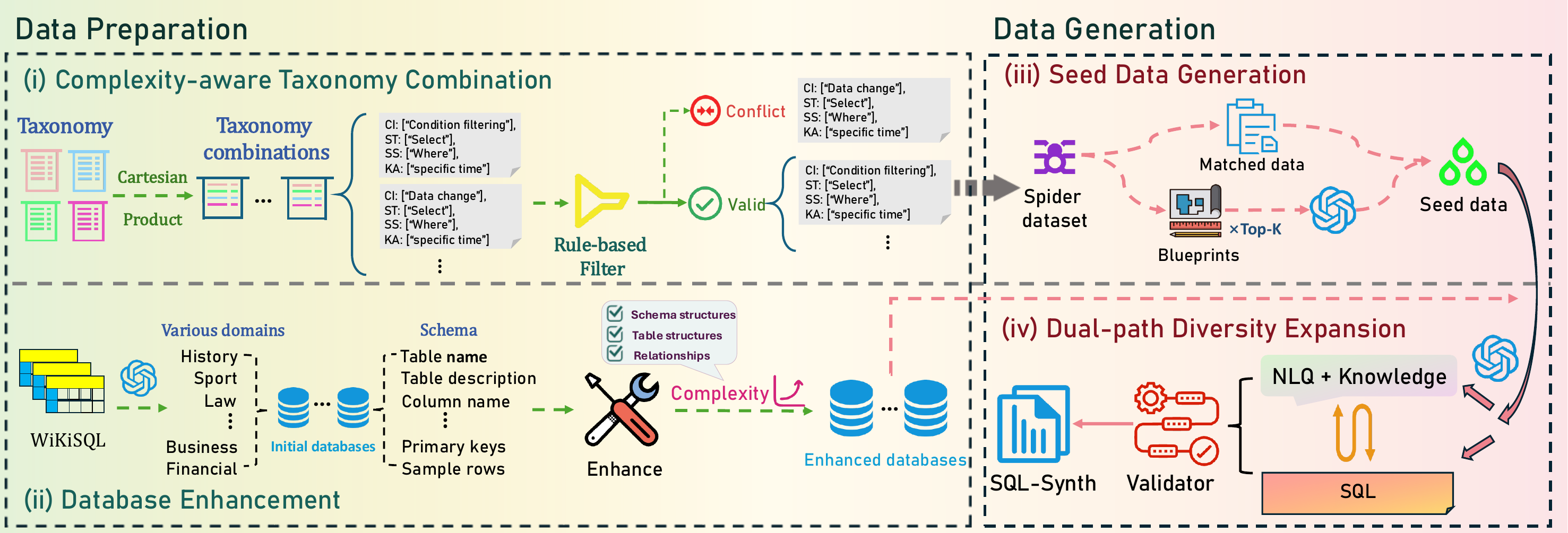}
  \caption{The taxonomy-guided dataset synthesis pipeline consists of four main processes: (i) Complexity-aware taxonomy combination, which generates valid taxonomy combinations under various complexity-level; (ii) Database enhancement to generate meaningful databases; (iii) Seed data generation, which produces high-quality seed data that serves as templates for diversity expansion; (iv) Dual-path diversity expansion, where LLMs expand the diversity of seed data leveraging the enhanced databases.}
  \label{fig:dataset}
  \vspace{-5mm}
\end{figure*}

\subsection{Synthesis Methodology}

The taxonomy-guided synthesis pipeline has four main components: Complexity-aware taxonomy combination, Database enhancement, Seed data generation, and Dual-path diversity expansion, as illustrated in Figure~\ref{fig:dataset} and Algorithm~\ref{alg:pipeline}. Its goal is to synthesize a dataset that covers all dimensions of the taxonomy while ensuring sufficient diversity. To achieve this, we begin with generating valid taxonomy combinations and enhancing databases based on WiKiSQL to guide the subsequent synthesis process. Next, we extract data from the Spider dataset that aligns with these combinations to serve as seed data. For combinations that remain uncovered, we leverage LLMs to generate user questions and SQL queries based on randomly sampled databases. Finally, to ensure diversity for each seed data, we expand it using different contexts and synthesis paths.

\begin{algorithm}[t]
\scriptsize
 \SetAlgoLined
 \SetKwFunction{NLQOriented}{\textsc{NLQOriented}}
 \SetKwFunction{SQLOriented}{\textsc{SQLOriented}}
 \SetKwFunction{GenerateSeed}{\textsc{GenerateSeed}}
 \SetKwFunction{InitializeDatabase}{\textsc{InitializeDatabase}}
 \SetKwFunction{Enhance}{\textsc{Enhance}}
 \SetKwFunction{GenerateDatabase}{\textsc{GenerateDatabase}}
 \SetKwFunction{Valid}{\textsc{Valid}}
 \SetKwFunction{Cartesian}{\textsc{Cartesian}}
 \SetKwFunction{Pipeline}{\textsc{Pipeline}}

 \SetKwInOut{KwIn}{Inputs}
 \SetKwInOut{KwOut}{Output}
 \KwIn{\scriptsize{Taxonomy $\mathcal{T}=\{ \mathcal{CT},\mathcal{ST},\mathcal{SS},\mathcal{KA} \}$; \\
 Table list in WiKiSQL $\mathcal{W}$; \\}}
 \KwOut{\scriptsize{Generated dataset $\mathcal{N}$}}
\SetKwProg{Fn}{Procedure}{:}{}
 \Fn{\Pipeline{$\mathcal{T}$, $\mathcal{W}$}}{
    \tcp{\color{blue} Complexity-Aware Taxonomy Combination}
    
     $\mathcal{C'}$ $\gets$ $\texttt{[]}$, $\mathcal{C}$ $\gets$ \Cartesian{$\mathcal{T}$}
    
    \For{\textup{\textbf{each} $(ci, st, ss, ka) \in \mathcal{C}$}}{
        \tcp{\color{purple} Rule-based filtering}
        \If{\Valid{$(ci, st, ss, ka)$}}{
            $\mathcal{C'}$.append($(ci, st, ss, ka)$)
        }
    }

    \tcp{\color{blue} Database Enhancement}

    $\mathcal{D}$ $\gets$ $\texttt{[]}$
    
    \For{\textup{\textbf{each} $\texttt{w} \in \mathcal{W}$}}{
        \tcp{\color{purple} Database Generation}
        $\mathcal{D}_{\text{i}} \gets $ \GenerateDatabase{$\texttt{w}$}

        \tcp{\color{purple} Database Enhancement}
        $\mathcal{D}'_{\text{i}} \gets $ \Enhance{$\mathcal{D}_{\text{i}}$}

        \tcp{\color{purple} Database Initialization}
        $\hat{\mathcal{D}}_{\text{i}} \gets $ \InitializeDatabase{$\mathcal{D}'_{\text{i}}$}

        $\mathcal{D}$.append($\hat{\mathcal{D}}_{\text{i}}$)
    }

    \tcp{\color{blue} Seed Generation}
    
    $\mathcal{S} \gets $ \GenerateSeed{$\mathcal{D}$, $\mathcal{C'}$}

    \tcp{\color{blue} Dual-path Diversity Expansion}
    
    $\mathcal{N} \gets \texttt{[]}$
    
    \For{\textup{\textbf{each} $(s,d) \in (\mathcal{S}, \mathcal{D})$}}{
        \tcp{\color{purple} SQL-to-Text}
        $\mathcal{N}_{\text{i}}$ $\gets$ \SQLOriented{$s$, $d$}

        \tcp{\color{purple} Text-to-SQL}
        $\mathcal{N}_{\text{i+1}}$ $\gets$ \NLQOriented{$s$, $d$}

        $\mathcal{N}$.append($\mathcal{N}_{\text{i}},\mathcal{N}_{\text{i+1}}$)

    }
    \KwRet $\mathcal{N}$
  }
 \caption{Taxonomy-Guided Synthesis Pipeline}
 \label{alg:pipeline}
\end{algorithm}

\noindent\textbf{Complexity-aware taxonomy combination} 
Building effective text-to-SQL systems requires comprehensive and representative datasets to ensure robustness and reliability. However, we observe from the real-world scenarios that existing datasets, such as Spider and Bird, often suffer from insufficient coverage across key dimensions. To address this limitation, we propose a new complexity-aware taxonomy combination method, leveraging our proposed taxonomy to ensure thorough coverage of all complexity levels and taxonomy dimensions.

To measure the complexity level of text-SQL pairs, we assign distinct scores to each dimension of the taxonomy based on their operational complexity and impact. Using these scores, we define three carefully designed complexity levels, namely simple, medium, and hard, each corresponding to a specific score range. We then generate all possible combinations of taxonomy through a Cartesian product, ensuring that the total complexity scores fall within the designated range. 
Furthermore, to resolve conflicts between certain dimensions across different taxonomies, we employ a rule-based method to remove invalid or conflicting combinations. This method effectively filters out problematic combinations, thereby improving the overall quality of the generated data. 
We define these combinations as $C=\{ct^i,st^i,ss^i,ka^i,cl^i\}_{i=1}^n$, where $i$ represents the $i$-th combination, $ct^i,st^i,ss^i,ka^i$ correspond to its core intent, statement type, syntax structure, and key action, $cl^i$ denotes its complexity level.

\noindent\textbf{Database enhancement} Developing robust text-to-SQL models necessitates fine-tuning on a diverse range of databases. However, we observe that existing datasets, such as Spider and Bird, contain a limited number of databases, which constrains their ability to support large-scale data synthesis. To address this limitation, we propose a key component in our pipeline that consists of two critical steps: database generation and database enhancement.

Specifically, we construct enhanced databases using WiKiSQL, which provides tables spanning various domains. For each table, we first prompt LLMs to design a realistic business scenario relevant to the table and then generate an initial database. Each initial database consists of multiple relational tables, complete with structural information such as primary keys and foreign keys. The schema for each table includes the table name, a description, column names, column descriptions, data types, and several sample rows.

However, due to the output length limitations of LLMs, the initial databases often suffer from overly simplistic table designs and database structures. 
To overcome these issues, we leveraging LLMs to enhance the complexity of the databases and enrich their relationships. This is achieved by adding new columns and refining foreign key connections, resulting in more robust and realistic database structures. 
After enhancing database schemas, we further initialize databases to obtain databases files. Please refer to Appendix~\ref{App:Database Initialization Algorithm} for more details.

\noindent\textbf{Seed data generation}
Given that providing LLMs with templates to replace key values results in more accurate and higher-quality text-SQL pairs compared to having them generate pairs entirely on their own, this component focuses on generating an initial set of seed text-SQL pairs to facilitate data synthesis.

Specifically, we begin by extracting data from the Spider train dataset that aligns completely with any combination. For each uncovered combination, we first retrieve the top-$K$ ($K=5$) most relevant blueprint data from Spider based on the Jaccard similarity~\citep{bag2019efficient,niwattanakul2013using,zahrotun2016comparison}. Next, we randomly sample a database schema to serve as context, which is then used to guide LLMs to modify the blueprints. We define the generated seed data as $S=\{q_i,s_i,C_i\}_{i=1}^n$, where $i$ represents the $i$-th text-SQL pair, $q_i,s_i,C_i$ correspond to its natural language question (NLQ), SQL query, and combination.

During the post-processing stage, we execute all seed data to identify those with syntax errors or those that result in timeouts. Then we instruct LLMs to correct these queries, ensurding the quality and reliability of the generated seed data.

\noindent\textbf{Dual-path diversity expansion}
After generating the seed data, the next step is to expand its diversity across various contexts. While existing methods primarily focus on enhancing dataset diversity through SQL queries, we argue that accurately capturing user intent is crucial for generating high-quality SQL queries. Our analysis reveals that SQL queries tend to abstract and obscure the origin user intents. Consequently, traditional synthesis methods \citep{QmniSQL, guo2025sqlforge, zhang2023sciencebenchmark} that prioritize generating SQL queries first, followed by corresponding user questions, fail to adequately represent common user intents encountered in real-world scenarios. 

To enhance data diversity, we propose an innovative \textit{dual-path} expansion method, consisting of SQL-oriented generation and question-oriented generation. The SQL-oriented generation focuses on syntax correctness and logic complexity. For each SQL query in the seed data, we randomly sample 50 databases as various context and instruct LLMs to generate new SQL queries, followed by corresponding user questions. On the other hand, the question-oriented generation emphasizes capturing user's core intents to produce diverse questions. For each user question in the seed data, we similarly sample 50 databases as well and instruct LLMs to generate new questions that aligns with the given databases, followed by corresponding SQL queries. 

Furthermore, to address the challenge posed by ambiguous user questions, which often makes it challenging for LLMs to accurately identify relevant tables or columns in a database, we augment the data by extracting external knowledge from both the user question itself and the LLM's knowledge base. 
Specifically, we categorize external knowledge into two types: value-mapping and numerical calculation. For the value-mapping, the LLM is instructed to identify specific values present in database schema or content, and replace them with clear and common descriptions. For the numerical calculation, the LLM is instructed to provide explanations for calculations only when the values require multi-step calculations.

To ensure the quality of the generated data, we utilize an execution validator and a semantic validator to filter out incorrect outputs, as well as text-SQL pairs with inconsistent semantics.

\subsection{Dataset Composition and Statistics}
\begin{table*}[htb]
    \centering
    \small
    \begin{tabular}{l|ccccccc}
        \toprule
        Dataset & Source & DB & Knowledge & \makecell{*Core\\Intent ($\uparrow$)} & \makecell{*Statement\\Type ($\uparrow$)} & \makecell{*Syntax\\Structure ($\uparrow$)} & \makecell{*Key\\Action ($\uparrow$)}\\ \midrule
        Spider & Human & 200 & \textcolor{black}{\ding{55}} & 0.79 & 0.2 & 0.71 & 0.33\\
        Bird & Human & 95 & \textcolor{red}{\ding{51}} & 0.86 & 0.2 & 0.57 & 0.89\\ 
        \midrule
        \method{} & LLM-Gen & 1250 & \textcolor{red}{\ding{51}} & 1 & 1 & 1 & 1\\ 
        \bottomrule
    \end{tabular}
    \caption{Comparison of our dataset with existing datasets in the text-to-SQL Task. An asterisk (*) indicates complete coverage of all dimensions.}
    \label{tab:Overall statistics of different datasets}
    \vspace{-5mm}
\end{table*}

Table~\ref{tab:Overall statistics of different datasets} shows a comprehensive comparison of \method{} with other text-to-SQL datasets. As demonstrated, \method{} is the first large-scale text-to-SQL dataset extensively grounded in real-world scenarios, encompassing a diverse range of domains and databases. Unlike Spider and Bird, \method{} is generated through an LLM-driven pipeline, ensuring both scalability and efficiency. Moreover, \method{} comprehensively covers the entire taxonomy. For ambiguous or metaphorical queries, external knowledge is integrated to clarify user intent.

\section{Experimental Evaluations}
In this section, we comprehensively evaluate the performance of
\method{} by comparing it with other leading LLMs.

\subsection{Experimental Setup}
\noindent \textbf{Datasets.} 
We generate training and test sets respectively using different database schemas with a ratio of around 10: 1. Evaluations were conducted on the test set.

\noindent \textbf{Evaluation metrics.} Following previous work, we use execution accuracy (EX) as the evaluation metric, which measures whether the predicted SQL query produces the same execution results as the corresponding gold SQL query. 

\noindent \textbf{Evaluation Models.} We evaluate our dataset with a wide range of LLMs, including cloased-source models such as GPT-4o-mini, GPT-4o, and GPT-4-Turbo, as well as open-source models like DeepSeek-V3~\citep{liu2024deepseek}, Qwen3 Coder~\citep{yang2025qwen3technicalreport}, Qwen2.5~\citep{qwen2025qwen25technicalreport}, Qwen2.5 Coder~\citep{hui2024qwen25codertechnicalreport}, Granite3.1~\citep{mishra2024granitecodemodelsfamily}, and our fine-tuned model Synth-Coder based on Qwen2.5-Coder-7B-Instruct. 

Please refer to the Appendix~\ref{App: Experimental Setup} for more detailed information.

\subsection{Performance Evaluation}
\begin{table}[t!]
    \centering
    \small
    \begin{tabular}{l|c}
        \toprule
        Model & \method{} (Test)\\ \midrule
        \multicolumn{2}{c}{Closed-source LLMs} \\ \midrule
        GPT-4o-mini & 79.57\\
        GPT-4-turbo & 81.24 \\
        GPT-4o & 85.05\\ \midrule
        \multicolumn{2}{c}{Open-source LLMs} \\ \midrule
        Qwen3-Coder & 82.63 \\
        Qwen2.5-32B-Instruct & 79.69 \\
        DeepSeek-V3 & 80.49 \\
        Granite-3.1-8B-Instruct & 68.18 \\
        Qwen2.5-7B-Instruct & 72.27 \\
        Qwen2.5-Coder-7B-Instruct & 74.52 \\ \midrule
        Synth-Coder & \textbf{85.12} \\ \bottomrule
    \end{tabular}
    \caption{Evaluations on our \method{}.}
    \label{tab:evaluations-main}
    \vspace{-5mm}
\end{table}

The evaluations on our \method{} test set are shown in the Table~\ref{tab:evaluations-main}. We compare Synth-Coder with open-source and close-source LLMs of different sizes. 

\noindent \textbf{Synth-SQL significantly enhances the base model’s text-to-SQL capabilities.} Fine-tuning with \method{} shows clear benefits, as evidenced by the comparison between Qwen2.5-Coder-7B-Instruct and Synth-Coder. Notably, Synth-Coder achieves a remarkable 14\% improvement over its base model, Qwen2.5-Coder-7B-Instruct. What's more, compared with much larger models including GPT-4o and DeepSeek v3, Synth-Coder also shows the best performance.

\begin{figure}[t!]
  \centering
  \includegraphics[width=0.45\textwidth]{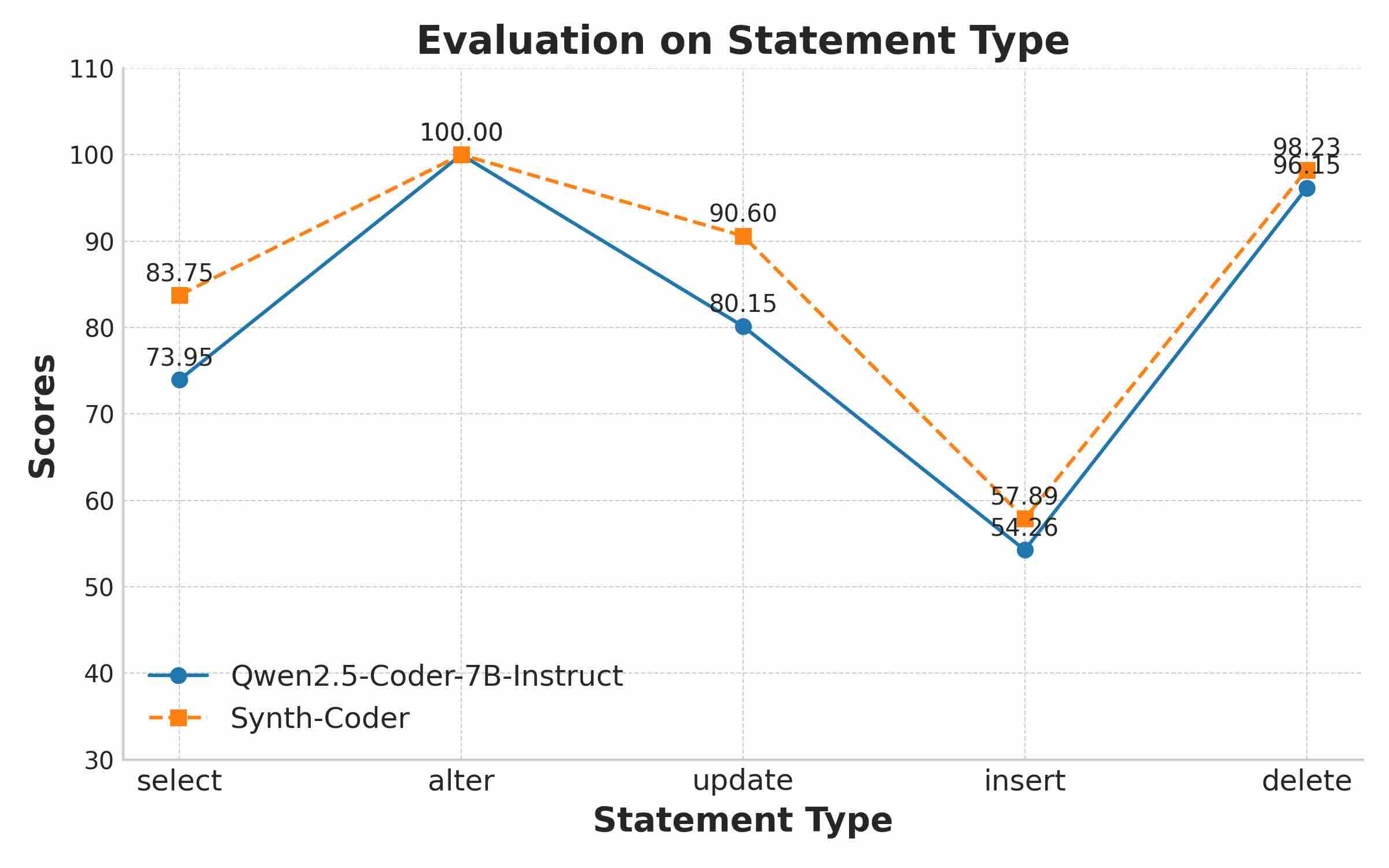}
  \caption{Evaluations on statement type.}
  \label{tab:evaluations-1}
  \vspace{-3mm}
\end{figure}

\noindent \textbf{Synth-SQL significantly enhances the model’s comprehension of each taxonomy.} Noteably, on our proposed taxonomy, Synth-Coder consistently delivers superior performance across the board. The evaluation results for statement type of our taxonomy are presented in the Figure~\ref{tab:evaluations-1}. As observed, Synth-Coder demonstrates an average improvement of 8.8\% across the five proposed statement types: SELECT, ALTER, UPDATE, INSERT, and DELETE. Specifically, for the most common statement type, SELECT, Synth-Coder achieves a remarkable 13.3\% improvement over the base model, revealing its effectiveness and generalizability. More evaluation results of our taxonomy are shown in the Appendix~\ref{Taxonomy Evaluations}.

\subsection{Data Quality Evaluation}

\begin{figure}[t!]
  \centering
  \includegraphics[width=0.45\textwidth]{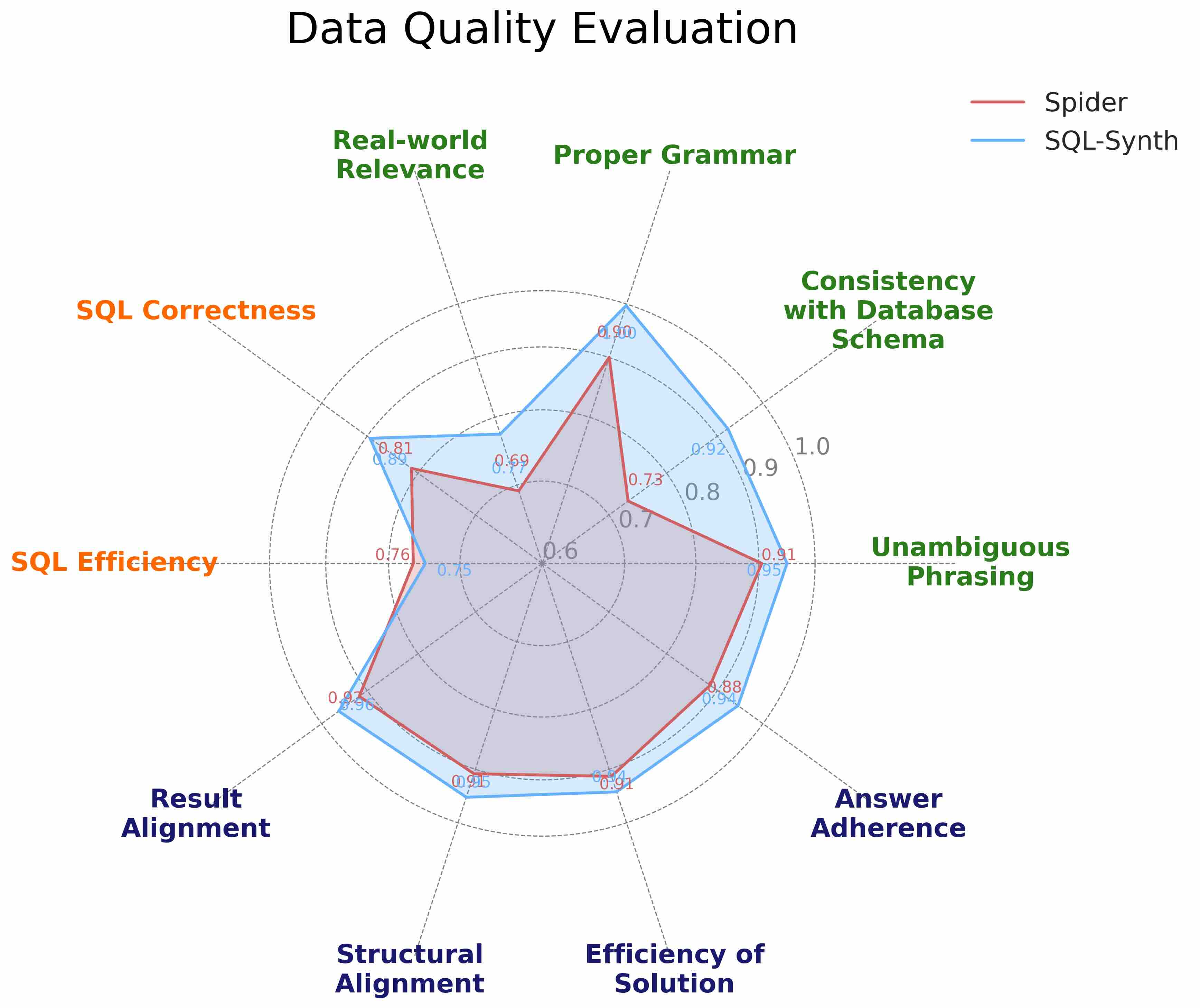}
  \caption{Quality evaluation of \method{} and Spider judged by GPT-4o. Scores are computed using Equation~\ref{score equation}.}
  \label{fig:data_quality_radar}
  \vspace{-3mm}
\end{figure}

In this section, we evaluate \method{} using GPT-4o, a state-of-the-art LLM, following the paradigm of recent LLM-as-a-judge studies~\citep{gu2025surveyllmasajudge, zhu2025judgelmfinetunedlargelanguage, QmniSQL}. The evaluations focuses on three key dimensions: \textbf{Question Aspect} (Real-world Relevance, Proper Grammar, Consistency with Database Schema, and Unambiguous Phrasing), \textbf{SQL Aspect} (SQL Correctness and SQL Efficiency), and \textbf{Result Aspect} (Result Alignment, Structural Alignment, Efficiency of Solution, and Answer Adherence). We instruct GPT-4o to assign one of four levels (Excellent, Good, Average, and Poor) for each criterion on each data sample, along with detailed explanations. We calculate the final scores using the following equation: 
\begin{equation}
    \small
    \label{score equation}
    Score=\frac{N_e * 1 + N_g * 0.75 + N_a + 0.5 + N_p * 0.25}{N_e + N_g + N_a + N_p},
\end{equation}
\noindent where $N_e$, $N_g$, $N_a$, and $N_p$ are the numbers of samples assigned with Excellent, Good, Average, and Poor. For comparison, we also conduct evaluations on Spider, a widely-used text-to-SQL dataset. For efficiency and fairness, we randomly sample 2000 samples for each dataset and results are shown in the Figure~\ref{fig:data_quality_radar}.

The results demonstrate that the quality of \method{} generally surpasses that of Spider, despite Spider being meticulously hand-crafted. However, it is worth noting that the efficiency of the generated SQL is lower. This is primarily due to the incorporation of taxonomy combinations, which can hinder the exploration and generation of the most efficient solutions. Nonetheless, the findings suggest that the majority of \method{} are high-quality and well-suited for model training.

\section{Conclusion}

This paper presents a novel taxonomy and pipeline for text-to-SQL analysis and synthesis. The proposed taxonomy encompasses four key dimensions, focusing on both the user question and the corresponding SQL query. Building on this taxonomy, we introduce an innovative data synthesis pipeline capable of generating high-coverage and diverse datasets. Leveraging this pipeline, we create \method{}, a new text-to-SQL dataset comprising over 100,000 high-quality samples.
Additionally, comprehensive evaluations of \method{} demonstrate its superior quality and robustness. Using this dataset, we develop Synth-Coder by fine-tuning Qwen2.5-Coder-7B-Instruct. We then benchmark Synth-Coder against nine baseline models, revealing that it achieves state-of-the-art performance while utilizing fewer parameters.

\section{Limitations}
We propose a taxonomy and a taxonomy-guided dataset synthesis pipeline, which effectively addresses the coverage and diversity challenges in dataset synthesis by integrating valid taxonomy combinations with enhanced databases. Furthermore, we utilize seed data and a dual-path expansion strategy to improve the quality of the data generated by the pipeline. 
However, our research in taxonomy is limited to open scenarios and domains, neglecting the requirements of closed-resource environments. Additionally, due to computational resources constraints, our fine-tuning experiments are restricted to LLMs with 7B parameters, leaving the performance of LLMs with other sizes unexplored. This limits our ability to fully analyze the effectiveness of the proposed taxonomy across different model scales. 
Moreover, the pipeline incurs high computational costs when applied to LLMs. To generate high-quality data, we instruct LLMs to enhance the database, produce seed data, and expand diversity in a step-by-step manner, which significantly increases token overhead. 
Therefore, future research could address these limitations by expanding the taxonomy to specific domains with diverse categories and template, exploring the performance of LLMs across different sizes to better understanding their capability, and investigate alternative LLM-driven pipelines that are more cost-effective and deliver higher performance.

\bibliography{ref}
\bibliographystyle{acl_natbib}

\onecolumn
\newpage

\twocolumn
\appendix

\section{Appendix}
\label{sec:appendix}

\subsection{Database Initialization Algorithm}\label{App:Database Initialization Algorithm}

\begin{algorithm}[th!]
\small
 \SetAlgoLined
 \SetKwFunction{UpdateInDegree}{\textsc{UpdateInDegree}}
 \SetKwFunction{CreateTable}{\textsc{CreateTable}}
 \SetKwFunction{GetZeroDegreeTables}{\textsc{GetZeroDegreeTables}}
 \SetKwFunction{ComputeInDegree}{\textsc{ComputeInDegree}}
 \SetKwFunction{DAGGeneration}{\textsc{DAGGeneration}}

 \SetKwInOut{KwIn}{Inputs}
 \SetKwInOut{KwOut}{Output}
 \KwIn{\small{Database schema list $\mathcal{D}$; \\}}
 \KwOut{\small{Created database list $\hat{\mathcal{D}}$}}
\SetKwProg{Fn}{Procedure}{:}{}
 \Fn{\Pipeline{$\mathcal{D}$}}{

    $\hat{\mathcal{D}}$ $\gets$ \text{[]}
    
    \For{\textup{\textbf{each} $\mathcal{D}_\text{i} \in \mathcal{D}$}}{
        $\mathcal{G}_\text{i}$ $\gets$ \DAGGeneration{$\mathcal{D}_\text{i}$}

        $\mathcal{I}_\text{i}$ $\gets$ \ComputeInDegree{$\mathcal{G}_\text{i}$}

        $\mathcal{Q}_\text{i}$ $\gets$ \GetZeroDegreeTables{$\mathcal{I}_\text{i}$}




        \For{\textup{\textbf{each} $\mathcal{M} \in \mathcal{Q}_\text{i}$}}{
            $\hat{\mathcal{D}}_\text{i}$ $\gets$ \CreateTable{$\mathcal{M}$}

            \UpdateInDegree{$\mathcal{I}_\text{i}$}

            $\mathcal{Q}_\text{i}$.append(\GetZeroDegreeTables{$\mathcal{I}_\text{i}$})
        }

        $\hat{\mathcal{D}}$.append($\hat{\mathcal{D}}_\text{i}$)
    }
    \KwRet $\hat{\mathcal{D}}$
  }
 \caption{Database Initialization}
 \label{alg:topological}
\end{algorithm}

After generating enhanced database schemas, the next step is to initialize them to create database files. A key challenge here is that the order of the generated tables may not align with their foreign key dependencies, potentially causing failures during initialization. 
Therefore, we utilize a topological sorting algorithm to address this issue, which is an algorithm designed to order the vertices of a directed acyclic graph (DAG) in a linear sequence that follows the direction of edges. 
As shown in Algorithm~\ref{alg:topological}, in our context, We model the foreign key dependencies between tables as a DAG and apply topological sorting algorithm to determine the correct order of table generation, ensuring the foreign key constraints satisfied.

\subsection{Experimental Setup}\label{App: Experimental Setup}

\begin{table*}[htb!]
    \centering
    \resizebox{1\textwidth}{!}{
    \begin{tabular}{l|cccccccccc}
        \toprule
        Dataset & DB & SQL & \makecell{Tables\\per SQL} & \makecell{Tokens\\per SQL} & \makecell{Func\\per SQL} & Join & \makecell{Window\\Func} & CTEs & Subqueries & Func\\ 
        \midrule
        \makecell{Synth-train} & 1250 & 114029 & 1.43 & 23.59 & 1.18 & 32772 & 3974 & 5812 & 786 & 134752 \\
        \makecell{Synth-test} & 500 & 8601 & 1.45 & 23.77 & 1.18 & 2585 & 291 & 437 & 55 & 10122\\ \bottomrule
    \end{tabular}
    }
    
    \caption{Dataset statistics overview.}
    \label{Dataset statistics}
\end{table*}

In the Table~\ref{Dataset statistics}, we present the statistics of the generated dataset using our pipeline focusing on databases number, samples number, average number of tables, tokens, and functions, and some key dimensions of taxonomy. It shows that \method{} is complex enough to reflect real-world scenarios. 

The approach for the calculation of the evaluation metric EX varies between different types of SQL statements. For SELECT statement, we directly compare the execution results of the predicted query and label query. For other statements, such as DELETE, ALTER, UPDATE, and INSERT, we evaluate by comparing the database state after the query is executed. 

For the main evaluation, Qwen2.5-Coder-7B-Instruct, a series of advanced code language model pre-trained and instruct-tuned on 92 programming languages, is chosen as the base model for fine-tuning using low-rank adaption (LoRA), obtaining Synth-Coder. For LoRA, we set $r=128$ and $\alpha=256$, integrating adapters into q\_proj, k\_proj, v\_proj, o\_proj, gate\_proj, up\_proj, and down\_proj of the model. We use a learning rate cosine scheduler with a linear warmup for the initial 5\% of training, followed by a peak rate $5e^{-5}$. In addition, the batch size, num of epochs, and max sequence length are set to 32, 3, and 3072, respectively. 

Specifically, all experiments were conducted using a single Nvidia A100-SXM-80GB GPU and an Intel Xeon Platinum 8336C CPU. For training and inference of LLMs, we utilized PyTorch 2.7.1, PEFT 0.15.2, and the Transformers library version 4.52.4. 
All results are based on single-run experiments.

\subsection{Taxonomy Evaluations}\label{Taxonomy Evaluations}
In this section, we present the evaluation results for core intents, syntax structures, and key actions to comprehensively demonstrate effectiveness and generalizability of our taxonomy. 
The results are shown from Table~\ref{tab:evaluations-4} to Table~\ref{tab:evaluations-2}. 

For core intents, it can be observed that LLMs struggle to accurately capture user intents for complex and rare scenarios, such as format transformation, distribution, business calculation and business rule, which often leads to incorrect SQL queries when translating user questions. After fine-tuning, the model's performance on these challenging intents are significantly improved with a maximum accuracy increase of 43.5\%. 
Similarily for syntax structures and key actions, LLMs performance on challenging dimensions are significantly improved after fine-tuning. 

\begin{table*}[htb]
    \centering
    \resizebox{1\textwidth}{!}{
        \begin{tabular}{l|ccccccc}
            \toprule
            Model & \makecell{Basic\\query} & \makecell{Condition\\filtering} & \makecell{Sorting\\and pagination} & \makecell{Basic\\aggregation} & \makecell{Time\\operation} & \makecell{Format\\transformation} & \makecell{Set\\operation} \\ \midrule
            
             Qwen2.5-Coder-7B-Instruct & 68.91 & 77.98 & 76.93 & 74.05 & 76.56 & 63.49 & 69.08 \\
             Synth-Coder & \textbf{80.06} & \textbf{86.12} & \textbf{86.72} & \textbf{84.86} & \textbf{87.73} & \textbf{78.66} & \textbf{78.97}\\ \midrule

             Model & \makecell{Data\\change} & \makecell{Structure\\change} & \makecell{Distribution\\analysis} & \makecell{Advanced\\statistics} & \makecell{Trend\\analysis} & \makecell{Business\\calculation} & \makecell{Business\\rule} \\ \midrule

             Qwen2.5-Coder-7B-Instruct & 79.10 & 70.37 & 64.78 & 45.45 & 53.12 & 69.49 & 77.32 \\
             Synth-Coder & \textbf{87.27} & \textbf{71.43} & \textbf{75.95} & \textbf{65.22} & \textbf{64.29} & \textbf{81.03} & \textbf{85.72}\\ \bottomrule
        \end{tabular}
    }
    \caption{Evaluations on core intent.}
    \label{tab:evaluations-4}
\end{table*}

\begin{table*}[htb]
    \centering
    \resizebox{1\textwidth}{!}{
        \begin{tabular}{l|ccccccc}
            \toprule
            Model & \makecell{Where} & \makecell{Order by} & \makecell{Limit offset} & \makecell{Inner  join} & \makecell{Cross join} & \makecell{Outer join} & \makecell{Group by} \\ \midrule
            
             Qwen2.5-Coder-7B-Instruct & 77.11 & 72.60 & 68.23 & 67.99 & 40.00 & 53.91 & 69.25 \\
             Synth-Coder & \textbf{85.59} & \textbf{82.90} & \textbf{79.53} & \textbf{75.45} & \textbf{60.00} & \textbf{69.26} & \textbf{81.80}\\ \midrule

             Model & \makecell{Having} & \makecell{Union} & \makecell{Intersect} & \makecell{Except} & \makecell{Scalar subquery} & \makecell{Correlated subquery} & \makecell{Common table expression} \\ \midrule
             Qwen2.5-Coder-7B-Instruct & 82.76 & 44.40 & 70.30 & 71.21 & 74.77 & 65.92 & 62.84\\
             Synth-Coder & \textbf{89.33} & \textbf{70.61} & \textbf{82.18} & \textbf{83.33} & \textbf{84.27} & \textbf{79.82} & \textbf{74.32}\\ \bottomrule
        \end{tabular}
    }
    \caption{Evaluations on syntax structure.}
    \label{tab:evaluations-3}
\end{table*}

\begin{table*}[htb]
    \centering
    \resizebox{1\textwidth}{!}{
        \begin{tabular}{l|ccccccccc}
            \toprule
            Model & \makecell{Specific\\time} & \makecell{Wildcard\\filterig} & \makecell{Time\\function} & \makecell{Json\\function} & \makecell{Aggregate\\function} & \makecell{Window\\function} & \makecell{String\\function} & \makecell{Cast} & \makecell{Condition\\judgement}\\ \midrule
            Qwen2.5-Coder-7B-Instruct & 81.93 & 81.50 & 72.95 & 20.00 & 74.48 & 55.16 & 75.28 & 72.62 & 57.19 \\
            Synth-Coder & \textbf{86.51} & \textbf{88.86} & \textbf{83.56} & \textbf{43.95} & \textbf{84.92} & \textbf{76.16} & \textbf{86.35} & \textbf{76.19} & \textbf{73.77} \\ \bottomrule
        \end{tabular}
    }
    \caption{Evaluations on key action.}
    \label{tab:evaluations-2}
\end{table*}

\subsection{Ablation Study on Dual-path}

\begin{table}[htb!]
    \centering
    \small
    \resizebox{0.45\textwidth}{!}{
        \begin{tabular}{l|c|c}
            \toprule
            Method & TTR & \makecell{Number of\\Semantic clusters}\\ \midrule
            \method{} w/o nlq2sql & 0.072 & 1159 \\ 
            \method{} & \textbf{0.097} & \textbf{1586} \\ \bottomrule
        \end{tabular}
    }
    \caption{Evaluations of data diversity.}
    \label{tab:data diversity}
\end{table}

\textbf{Diversity Priority} In this part, we assess the impact of NLQ-to-SQL path on data diversity. 
Specifically, we remove the NLQ-to-SQL path in the Dual-path Diversity Expansion to generate a new dataset. 
Then we randomly sample 2000 data points and utilize two metrics, namely Type-Token Ratio (TTR) and the Number of Semantic clusters, to quatify the diversity of data. TTR is defined as the ratio of the number of unique words to the total number of words in a given text, and Number of Semantic clusters is calculated using a community detection algorithm applied to the vector representations of the text. 
As shown in the Table~\ref{tab:data diversity}, removing the NLQ-to-SQL path has negative impacts on data diversity, resulting in a 25\% reduction in TTR and 27\% decrease in the Number of Semantic clusters. This demonstrates that generating the user question first significantly enhances data diversity.

\begin{figure}[thb!]
  \centering
  \includegraphics[width=0.45\textwidth]{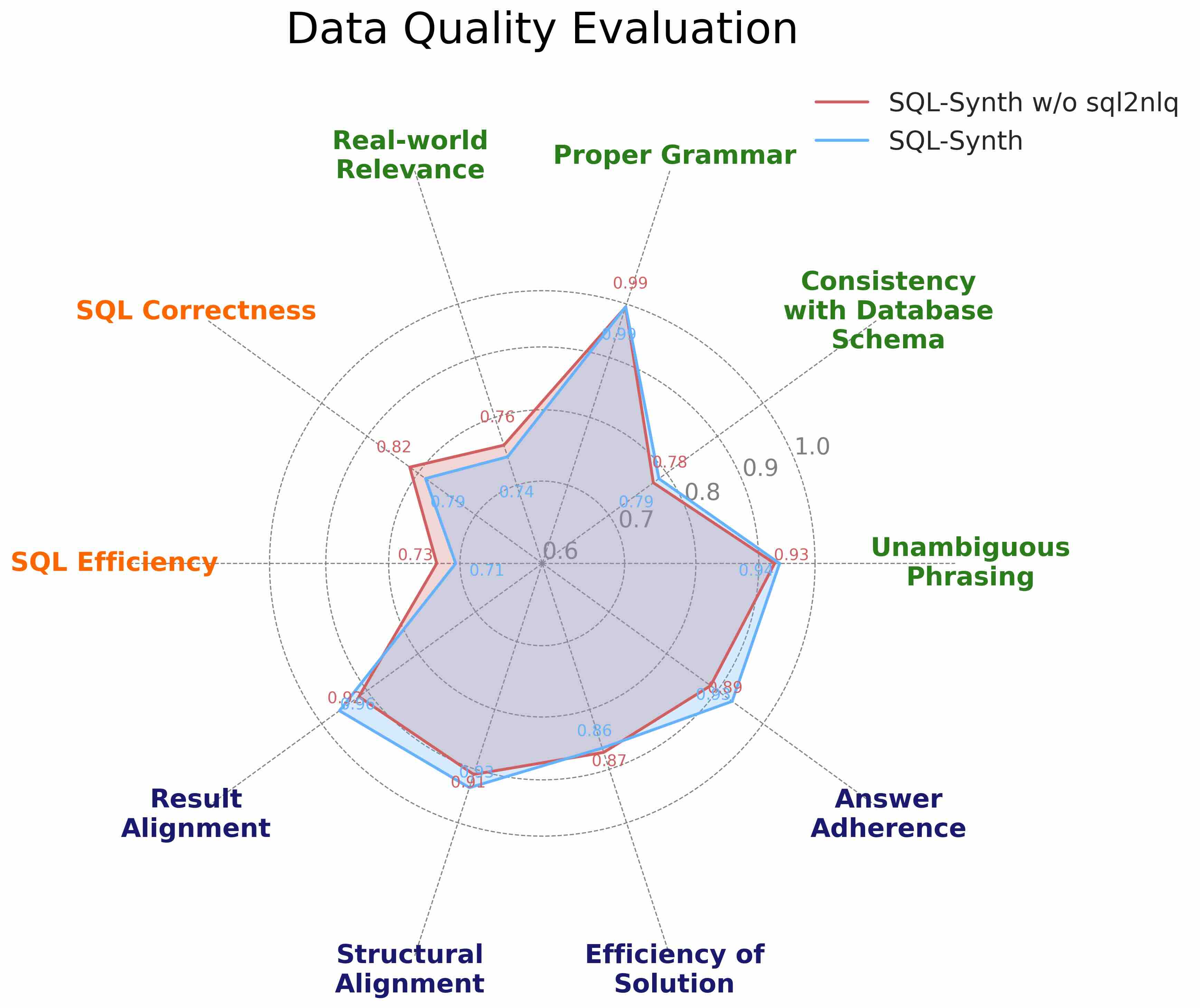}
  \caption{Quality evaluation of \method{} and \method{} w/o sql2nlq judged by GPT-4o. Scores are computed using Equation~\ref{score equation}.}
  \label{fig:ablation_data_quality_radar}
\end{figure}

\textbf{Quality Priority} In this part, we assess the impact of SQL-to-NLQ path on data quality. Specifically, we remove the SQL-to-NLQ path in the Dual-path Diversity Expansion to generate a new dataset. Then we randomly sample 2000 data points and evaluate data quality based on the Equation~\ref{score equation}. 
As observed from the Figure~\ref{fig:ablation_data_quality_radar}, the SQL efficiency and correctness are improved when utilizing only the NLQ-to-SQL path. This is primarily because the constraints imposed by the generated user question can limit the SQL generation process, encouraging it to focus on simpler and more efficient queries rather than handling complex and meaningful scenarios. Additionally, generating the user question first enhances real-world relevance, as the semantic of the user question are more closely aligned with real-world contexts. 
However, the result alignment, structural alignment, and answer adherence are significantly worse when relying solely on the NLQ-to-SQL path. This is mainly due to the openness and ambiguity inherent in the generated user question, which often leads to inaccurate or imprecise translations into SQL. This demonstrates that generating SQL first can significantly enhances data quality.

\subsection{Analysis on Training Data Scale Impact}

\begin{figure}[thb!]
  \centering
  \includegraphics[width=0.45\textwidth]{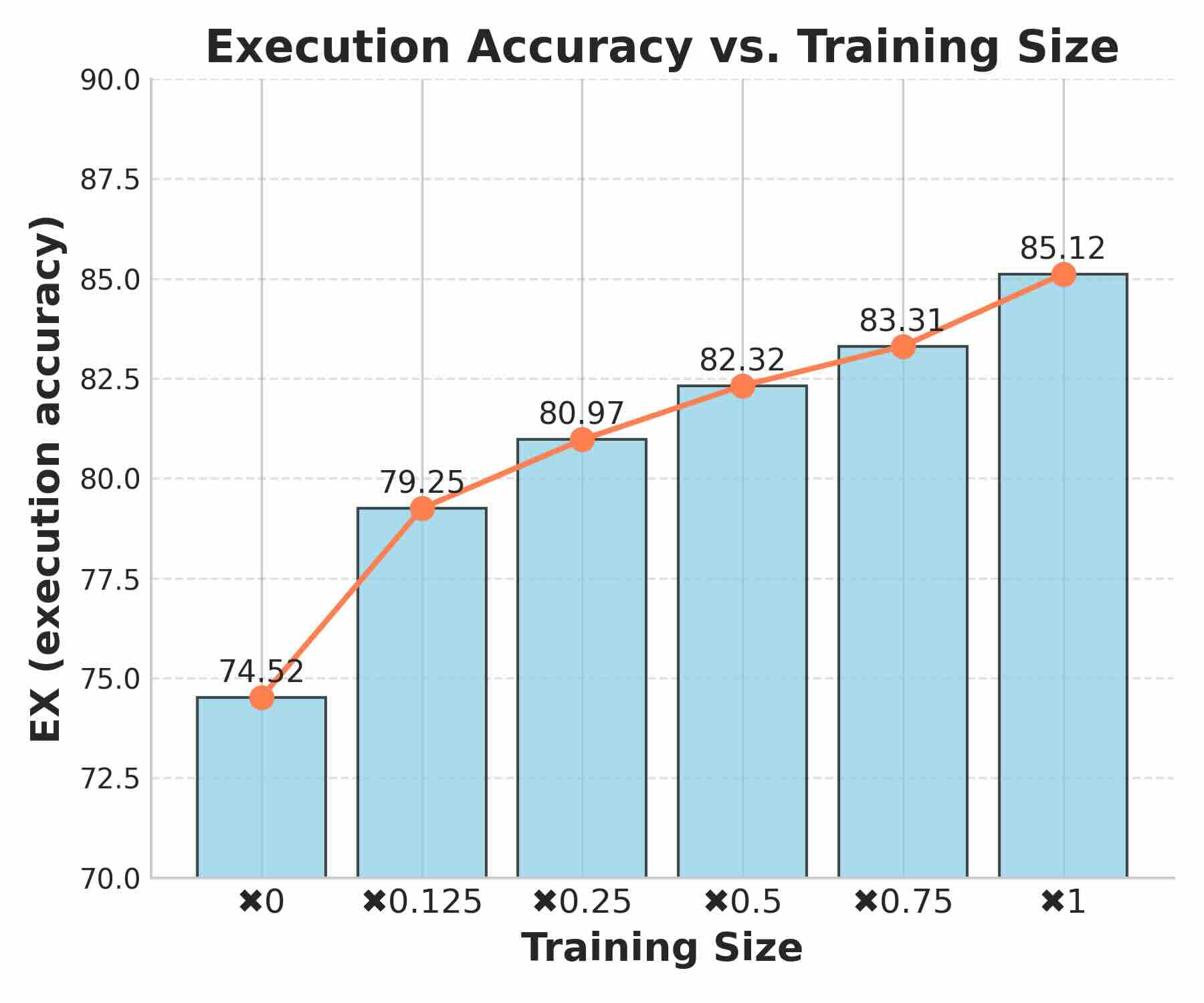}
  \caption{Effect of different scaling of the generated data
with Qwen2.5-Coder-7B-Instruct as the base model.}
  \label{fig:training size}
\end{figure}

Here we analyze the impact of scaling the generated data on model performance. Specifically, we train the base model on datasets scaled to $\times0$, $\times0.125$, $\times0.25$, $\times0.5$, $\times0.75$, and $\times1$, and the results are presented in the Figure~\ref{fig:training size}. 
It can be observed that as the scaling of the generated data increases, the model's performance consistently improve. This trend highlights the effectiveness of \method{} in leveraging larger amounts of generated data to enhance model training, enabling it to better capture task-specific patterns and improve generalization. Furthermore, this finding underscores the importance of data scalability in scenarios where real-world data is limited or expensive to obtain.

\subsection{Analysis on Seed Data Flexibility}

As mentioned before, the seed data component primarily serves to facilitate data synthesis, enabling the generation of more accurate and higher-quality data. We believe that the distribution of the seed data plays a crucial role in shaping the distribution of the final generated dataset. Consequently, by carefully designing seed data or constructing it from different datasets with varying distributions, we can effectively control the distribution of the generated dataset, thereby enhancing the flexibility of the pipeline. 
In this part, we utilize two generated dataset to train our base model, Qwen2.5-Coder-7B-Instruct, and evaluate their performance on the Spider dev and test dataset. 
Specifically, one dataset is generated using our pipeline as described earlier, while the other is created using a new seed dataset entirely synthesized by LLMs without leveraging any existing dataset. 

\begin{figure}[thb!]
  \centering
  \includegraphics[width=0.45\textwidth]{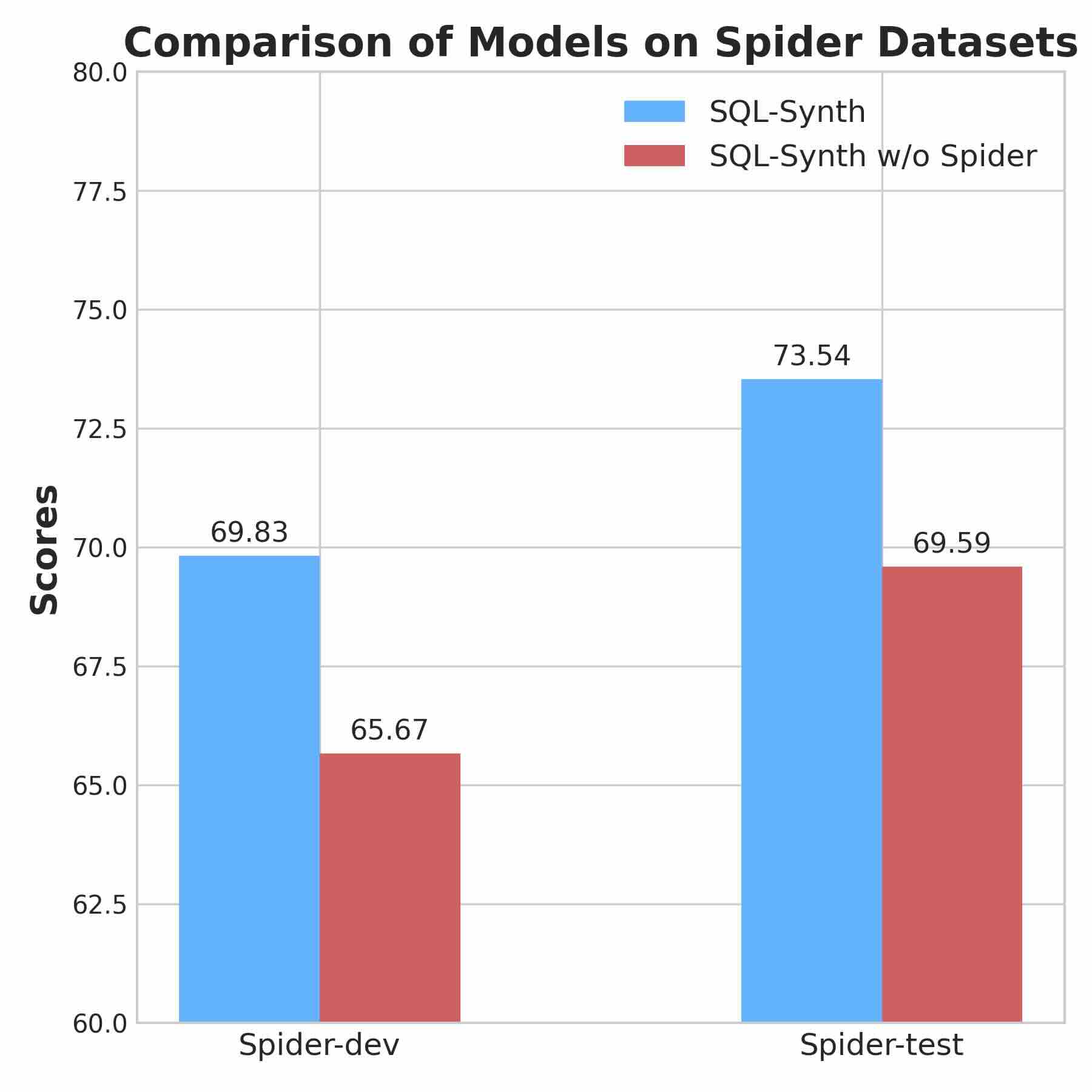}
  \caption{Comparison of Models trained on \method{} and \method{} w/o Spider.}
  \label{fig:flexibility}
\end{figure}

As illustrated in Figure~\ref{fig:flexibility}, the model trained using \method{} achieves a significant improvement compared to the model trained with \method{} without Spider. This suggests that the final dataset distribution can be effectively influenced by controlling the seed data distribution, thereby improving the model performance on specific scenarios and domains. Such an approach holds great potential for practical applications in real-world settings.

\subsection{Taxonomy Details}\label{Taxonomy Details}
In this section, we will introduce specific definitions about different dimensions within \textbf{Syntax structures} and \textbf{Key actions}.

\subsubsection{Syntax structures}
As discussed, syntax structures represent the rules and components used to construct SQL queries, which are decisive for performing specific operations. The detailed definitions of syntax structures are as follows:

\noindent\textbf{Where:} Present if the SQL includes WHERE followed by a conditional expression. 

\noindent\textbf{Order by:} Present if the SQL includes ORDER BY followed by column names, expressions, or ASC/DESC. 

\noindent\textbf{Limit offset:} Present if the SQL includes LIMIT to restrict the number of rows returned, optionally followed by OFFSET to skip a specified number of rows before starting to return rows.

\noindent\textbf{Inner join:} Present if the SQL uses INNER JOIN (or JOIN for short) with an ON clause.

\noindent\textbf{Cross join:} Present if the SQL uses CROSS JOIN or lists tables separated by commas without an ON clause.

\noindent\textbf{Outer join:} Present if the SQL uses LEFT OUTER JOIN, RIGHT OUTER JOIN, or FULL OUTER JOIN.

\noindent\textbf{Group by:} Present if the SQL includes GROUP BY followed by column names.

\noindent\textbf{Having:} Present if the SQL includes HAVING followed by a condition on aggregate values.

\noindent\textbf{Union:} Present if the SQL uses UNION to merge two result sets.

\noindent\textbf{Intersect:} Present if the SQL uses INTERSECT to return the common rows between two result sets. 

\noindent\textbf{Except:} Present if the SQL uses EXCEPT to return rows from the first result set that are not present in the second result set.

\noindent\textbf{Scalar subquery:} Present if the SQL includes a subquery in a scalar position.

\noindent\textbf{Correlated subquery:} Present if the subquery depends on outer query columns. 

\noindent\textbf{Common Table Expression:} Present if the SQL uses WITH to define one or more common table expressions (CTEs). 

\subsubsection{Key actions}
As discussed, key actions represent specific operations or functionalities performed within a SQL query to manipulate, filter, transform, or analyze data. The detailed definitions of key actions are as follows:

\noindent\textbf{Specific time:} Present if the SQL includes hardcoded temporal values.

\noindent\textbf{Wildcard filtering:} Present if the condition filters rows based on a column's match to a wildcard pattern using LIKE.

\noindent\textbf{Time function:} Present if the SQL uses any function that operates on temporal data.

\noindent\textbf{Json function:} Present if the SQL interacts with JSON data using functions.
 
\noindent\textbf{Window function:} Present if the SQL includes any function followed by OVER().

\noindent\textbf{String function:} Present if the SQL uses these non-regex string operations.

\noindent\textbf{Cast:} Present if the SQL uses CAST function to convert data types, excluding implicit conversions.

\noindent\textbf{Condition judgment:} Present if the SQL uses conditional logic to determine output values.

\noindent\textbf{Aggregate function:} Present if the SQL uses multi-row calculation functions.

\subsection{Prompt Design}

Prompt~\ref{prompt:database generation} and Prompt~\ref{prompt:database enhancement} represent the prompts used in the Database enhancement.
Prompt~\ref{Prompt of question generation}, Prompt~\ref{Prompt of knowledge generation}, and Prompt~\ref{Prompt of sql generation} represent the prompts used in the Dual-path diversity expansion.

\onecolumn

\begin{figure*}[t]
\begin{lstlisting}[breaklines=true,label={prompt:database generation},caption={Prompt for the \textit{database generation}.}]
**Task Overview**:  
Given a table's schema, your task is to complete the following two tasks:

## Task 1  
Analyze the table's header and rows to briefly summarize its domain.  
The output domain should be surrounded by <Domain> and </Domain>. 

## Task 2  
Based on the table, propose a complex, enterprise-level, real-world business scenario.  
The output scenario should be surrounded by <Scenario> and </Scenario>. 

Here, we present a demonstration for reference:  
{Example here...} 

Now, complete the tasks and propose a business scenario based on the table schema in the JSON format:  
<Table Schema>  
{schema}
\end{lstlisting}
\end{figure*}

\begin{figure*}[t]
\begin{lstlisting}[breaklines=true,label={prompt:database enhancement},caption={Prompt for the \textit{database enhancement}.}]
**Task Overview:**
Given a domain and a business scenario in this domain, your task is to complete the following task: 

## Task 
Create a comprehensive database schema in json format for the business scenario. The output database schema should be surrounded by <Generated Schema> and </Generated Schema>. 

The schema for each table should comprise the following seven key attributes: 
- table_name: The identifier or name assigned to the table. 
- table_description: A brief narrative explaining the purpose or content of the table. 
- column_names: A list of all column names within the table. Note that some column names might be abbreviations or non-descriptive strings, reflecting potential annotation inconsistencies in real-world database scenarios. 
- column_types: The data types assigned to each column, acknowledging that some types might be more complex (e.g., Date, Array, Struct) than basic data types. 
- column_descriptions: Detailed explanations for each column, providing semantic clarity on the data they hold. 
- primary_key: The column(s) designated as the primary key for the table, ensuring the uniqueness of each row. 
- sample_rows: An inclusion of two sample rows of data for the table, offering a practical glimpse into the table's data structure and content. 

Upon detailing all tables, establish and define the foreign key relationships within the database schema: 
- foreign_keys: A specification of the relationships between tables, indicating the source table, the referenced table, and the columns that are involved in this relationship. 

Here are the refined notes for Task 3: 
1. **Database Schema Design**: Develop a detailed and comprehensive set of tables and columns to accurately represent the complexity, scalability, and real-world business requirements of an enterprise-level application. 
2. **Table and Foreign Key Constraints**: Ensure that the specified number of tables are created, incorporating the appropriate number of foreign key constraints to maintain data integrity and relationships. 

Here, we present a demonstration for reference: 
{Example here...} 

Now, complete the tasks and generate a database schema containing {table_num} tables based on the following domain and scenario: 
{scenario}
\end{lstlisting}
\end{figure*}

\begin{figure*}[t]
\begin{lstlisting}[breaklines=true,label={Prompt of question generation},caption={Prompt for the \textit{question generation}.}]
**Task overview**
You are a data scientist specializing in the Text2SQL field. Your are provided with a testpoint combination, and a database schema. Your task is to understand the schema and create a high-quality natural language question/instruction that addresses real-world data analysis needs. 

<Instructions> 
- **Question/Instruction Generation** 
    - Ensure the generated question or instruction strictly align with the provided database schema, including column names, table names, and data types.
    - Ensure the generated question or instruction is clear and concise. 
    - Ensure the generated question or instruction addresses real-world data analysis needs. Avoid trivial or nonsensical query. 
    - Incorporate as many relevant tables and columns as possible into the generated question or instruction. 
    - Integrate the provided testpoint combination into the generated question or instruction where applicable, ensuring clarity, meaningfulness, and relevance. 
    - Use at least {table_num} tables into the generated question or instruction. 
    - Ensure the generated question or instruction obey the provided language style. 
</Instructions> 

<Language Style> 
{language_style} 
</Language Style> 

<Provided Database Schema> 
{schema} 
</Provided Database Schema> 

<Testpoint Combination> 
{tp_comb} 
</Testpoint Combination> 

**Output Format** 
Your output must be in json format that can be parsed with json.loads directly without any placeholders or explanations. 
{ 
    "question/instruction": "..." 
} 

Take a deep breath, think step by step, and then carefully create the question/instruction.
\end{lstlisting}
\end{figure*}

\begin{figure*}[t]
\begin{lstlisting}[breaklines=true,label={Prompt of knowledge generation},caption={Prompt for the \textit{knowledge generation}.}]
**Task overview** 
You are an expert in the Text2SQL field. Your are provided with a user question/instruction and database schema. Your task is to extract knowledge from the provided user question/instruction, and regenerate a high-quality question/instruction that replaces the extracted knowledge with common descriptions. 

<Instructions> 
- **External Knowledge Extraction** 
    - Analyze the user question/instruction and the database schema to identify any knowledge that is embedded in the question/instruction. 
    - Extract only the following types of knowledge into the "knowledge" field and replace the extracted knowledge with common descriptions in the question/instruction: 
        1. **Specific Value**: Replace specific values in the question/instruction with their real-world descriptions **only if the values are not common descriptions**. 
            - Example: 
                {Examples here ...} 
        2. **Numerical Calculation**: Explain logic for calculations only if the values are not directly present in the schema and require multistep calculations. 
            - Example: 
                {Example here ...} 
    - If no embedded knowledge is found, set "knowledge" to an empty string and retain the original question/instruction. 
- **Question/Instruction Regeneration** 
    - If knowledge is extracted, replace the extracted knowledge with common descriptions in the question/instruction to: 
        1. Exclude the extracted knowledge while preserving its original intent. 
        2. Be locally valid and concise. 
</Instructions> 

<Database Schema> 
{schema} 
</Database Schema> 

<User Question/Instruction> 
{question} 
</User Question/Instruction> 

**Output Format** 
Your output is supposed to be json format without any placeholders or explanations, which can be parsed with json.loads function. 
{ 
    "question/instruction": "..." 
    "knowledge": "..." or "" 
} 

Take a deep breath, think step by step, and then carefully extract knowledge before regenerating the question/instruction.
\end{lstlisting}
\end{figure*}

\begin{figure*}[t]
\begin{lstlisting}[breaklines=true,label={Prompt of sql generation},caption={Prompt for the \textit{sql generation}.}]
**Task overview** 
You are a data scientist specializing in the Text2SQL field. Your are provided with a database schema, user question/instruction, and external knowledge. Your task is to generate a executable SQL query to address the user question/instruction. 

<Instructions> 
- Ensure the generated SQL query is valid and executable in SQLite, and fully matches the database schema. 
- Ensure the generated SQL query accurately addresses the user question/instruction. 
- Incorporate the external knowledge to generate SQL query if provided. 
- Focus on generating a concise, clear, and efficient SQL query. 
</Instructions> 

<Database> 
SQLite 
</Database> 

<Database Schema> 
{schema} 
</Database Schema> 

<User Question/Instruction> 
{question} 
</User Question/Instruction> 

<External Knowledge> 
{knowledge} 
</External Knowledge> 

**Output Format** 
Your output is supposed to be json format without any placeholders or explanations, which can be parsed with json.loads function. 
{ 
    "sql": "...", 
} 

Take a deep breath and think step by step to generate an executable SQL query to address the user question/instruction.
\end{lstlisting}
\end{figure*}

\end{document}